\documentclass[11pt,a4paper]{article}
\usepackage[hyperref]{acl2021}
\usepackage{times}
\usepackage{latexsym}

\usepackage{graphicx}
\usepackage{caption}
\usepackage{subcaption}

\usepackage{microtype}
\usepackage{soul}

\aclfinalcopy 

\title{NLP for Climate Policy: Creating a Knowledge Platform for Holistic and Effective Climate Action} 

\author{Pradip Swarnakar$^{1}$ \qquad Ashutosh Modi$^{2}$ \\
  $^{1}$Department of Humanities and Social Sciences\\
  $^{2}$Department of Computer Science and Engineering\\
    Indian Institute of Technology Kanpur, Kanpur 208016, India \\
  $^{1}$\texttt{spradip@iitk.ac.in} \qquad $^{2}$\texttt{ashutoshm@cse.iitk.ac.in}
  }
  
\date{}

\begin{document}
\maketitle
\begin{abstract}

Climate change is a burning issue of our time, with the Sustainable Development Goal (SDG) 13 of the United Nations demanding global climate action. The worldwide scientific evidence exhibits that the current processes of market-led economic development are unsustainable. Realizing the urgency, in 2015 in Paris, world leaders signed an agreement committing to taking voluntary action to reduce carbon emissions. However, the scale, magnitude, and climate action processes vary globally, especially between developed and developing countries. Therefore, from parliament to social media, the debates and discussions on climate change gather data from wide-ranging sources essential to the policy design and implementation. The downside is that we do not currently have the mechanisms to pool the worldwide dispersed knowledge emerging from the structured and unstructured data sources.

The paper thematically discusses how NLP techniques could be employed in climate policy research and contribute to society's good at large. In particular, we exemplify symbiosis of NLP and Climate Policy Research via four methodologies. The first one deals with the major topics related to climate policy using automated content analysis. We investigate the opinions (sentiments) of major actors' narratives towards climate policy in the second methodology. The third technique explores the climate actors' beliefs towards pro or anti-climate orientation. Finally, we discuss developing a `Climate Knowledge Graph.'

The present theme paper further argues that creating a knowledge platform would help in the formulation of a holistic climate policy and effective climate action. Such a knowledge platform would integrate the policy actors' varied opinions from different social sectors like government, business, civil society, and the scientific community. The research outcome will add value to effective climate action because policymakers can make informed decisions by looking at the diverse public opinion on a comprehensive platform.

\end{abstract}

\section{Introduction}


Climate change is one of the gravest threats that humanity has ever faced. The natural systems essential for sustaining human and wildlife are fast depleting due to anthropogenic activities, the most dominant of which is the burning of fossil fuels \cite{ipcc2018}. The adverse environmental impacts of climate change spill over to interact with broader societal dynamics. The hierarchical power structures existing at the international and national levels determine how various social groups and individuals are affected by climate change. The poorest population of the world, although responsible for just 10 percent of global emissions \cite{oxfam2015}, bear the greatest brunt of climate change \cite{greenpeace-inequality} including climate-induced migration \cite{imo2020}. 

The international policy negotiations recognize the aforementioned socio-economic divide within the broader climate change concerns. The principle of Common but Differentiated Responsibility (CBDR) encapsulates this distinction recognizing that the “developed country Parties should take the lead in combating climate change and the adverse effects thereof” along with giving full consideration to “the specific needs and special circumstances of developing country Parties” \cite{unfcc1992}. 
From the climate change convention at Rio in 1992 to the Paris Agreement in 2015 \cite{paris2015}, this principle has been an overarching debating point between developed and developing countries while they negotiate on mitigating and adaptive actions like emissions reductions, energy transition to renewable sources, and climate funding for developing countries. The principle holds particular relevance for developing countries like India, which are trying to balance climate change action and ensure economic development to lift millions out of poverty. Moreover, United Nations Sustainable Development Goal (SDG) 13 \cite{sdg, sdgWeb, sdg13} needs to align with the Paris agreement because of holistic and urgent climate action \citep{campbell2018urgent}.

These debates around climate policy spill over into the public discourse, and international policy norms get domesticated through mass media like newspapers \citep{djerf2016framing} and social media sites\footnote{e.g., \url{https://twitter.com/Fridays4future}}. The mass media presents an arena where various climate-related claims get discussed and contested by diverse claim-makers, a process that has a bearing on public perception and policy \cite{boykoff2010indian}. Media discourse, therefore, offers “publicly visible sites for conflict over appropriate policy responses and approaches to climate governance.” \cite[p.402]{stoddart2015canadian}.   The use of language in a particular way, the negotiated choices of phrasing a problem “illustrate the acknowledged interaction between discourse and practice.” \cite [p. 11] {boykoff2011speaks}.

Given the importance of linguistic discourse in various media regarding climate change and related policy, analyzing the media discourse becomes pertinent. Not only it adds to our more extensive understanding of public perceptions on climate change, but it may also prove to be an effective tool at the disposal of policy-makers to make informed decisions, especially for developing countries like India, where reliable data will ease the humongous task of climate policy-making within the concerns of economic development.

Natural Language Processing (NLP) and its methodological toolbox provide a promising set of technical tools to make a robust analysis of linguistic discourse on climate change. The application of NLP techniques can help to reduce the tedious effort and biases of manual extraction. With this context in mind, this paper's overall objective is to discuss ways in which the NLP techniques can be exploited to collect, collate and systematically analyze the considerable amount of scattered data to render it useful for the purposes of research and policy-making. This will eventually lead to processing and gathering the text data faster, thus expediting the climate policy formulation process. 

There has been considerable research in linguistics and sociolinguistics in media discourse analysis \cite{o2011media, bednarek2006evaluation, fairclough1988discourse}. Researchers have investigated media discourse from different perspectives, e.g., genre analysis (linguistic analysis of newspaper genre and how it differs from other genres of a language) \cite{tardy2014genre}, register variation (modulation of language based on context and intended audience) \citep{enwiki:995093528, Lyons2013,biber1995dimensions}, conversational analysis (the study of the structure of conversations between individuals in any real-life setting) \citep{enwiki:1002133168, paulus2016applying}, critical discourse analysis (the study of discourse as a social practice) \citep{enwiki:1003585910, bisceglia2014combining, fairclough2001critical}, inter alia. Print media and social media texts have also been analyzed extensively in the NLP community via a number of tasks like opinion mining \citep{ravi2015survey}, controversy detection \citep{jang2016probabilistic, jang2016improving, dori2015controversy}, fake news detection \citep{zhou2020survey,yang2019unsupervised,thorne2017fake}, argument mining \citep{lippi2016argumentation}, stance detection \citep{ghosh2019stance, kuccuk2020stance}, perspective identification \citep{wong2016quantifying}, inter alia. Researchers have recently started exploring media's perception on climate issues \citep{jiang2017comparing, luo2020desmog}. 

However, to the best of our knowledge, research in the NLP domain on analyzing media discourse exclusively in the context of climate change policy has been very sparse. This line of research is still in its incipient stages, and a lot more is required to be done. Moreover, there exists a gap between the climate policy researchers and the NLP researchers. Both the communities have mostly been working in isolation. Climate policy researchers and climate policymakers, being from a different background, are unaware of different techniques proposed in the NLP community that could help glean and analyze pertinent information from myriad sources. On the other end of the spectrum, NLP researchers are unaware of the research and technical challenges pertinent to climate change research, hence are unable to contribute towards it. Both the communities need to talk to each other to create a significant impact.  

This paper aims to bridge the gap. We describe how techniques developed by the NLP community could contribute towards making a social impact. It explores how NLP techniques could be applied in the context of climate change policy and provide a means for the policy creators to make an informed decision. Though there could be myriad ways to address research in climate change policy analysis. This paper proposes four principal methodologies that could contribute to creating a knowledge platform that would help in the formulation of a holistic climate policy and effective climate action. The knowledge platform aims to integrate information from policy actors, the scientific community, government, and citizens. However, we firmly believe this is not an exhaustive list; there are many more exciting applications where NLP methodologies could be employed for climate change policy research. This paper argues that even relatively existing and straightforward NLP techniques can help bring about a significant social impact. We want to make the NLP community aware of the research avenues in climate change policy and instigate interest in this under-researched area. The proposed tasks serve as exemplars for the community, and we believe further research on these would spark new ideas. The methodologies of interest contributing towards the creation of knowledge platform are:
\begin{itemize}
\item Content Analysis of climate change policy via Topics.
\item Opinion Analysis of major actor's narratives.
\item Belief Analysis of major actors towards pro or anti-climate orientation.
\item Climate knowledge graph that links actors based on their beliefs. 
\end{itemize}

\section{Related Work}
\citet{huntingford2019machine} has argued that the integration of machine learning with rising concerns about climate change will provide abundant opportunities. This idea is evident by the increasing corpus of studies on climate change that are employing the toolbox of Machine learning to devise innovative ways to approach multifaceted challenges posed by warming earth and changing climate. The various techniques of machine learning have been employed for diverse uses, including Earth system modeling \citep{schneider2017earth,gentine2018could}, weather forecasting \citep{mcgovern2017using}, forecasting climate extremes like drought \citep{deo2015application}, among others.

The NLP techniques have been used to study the argumentative structure of the public debates. These are being employed in various studies, for example, for analyzing the global warming stances taken by various actors \citep{luo2020desmog}, uncovering the belief structure of actor-networks around climate change using Discourse Network Analyzer \cite{leifeld2017discourse, kukkonen2018international} and tracing coalitions in international climate negotiations \citep{castro2020national}. 

Researchers in communities other than NLP and ML have used the readily available tools to analyze print media for various goals, such as finding out relevant themes/topics, sentiments, and beliefs of the people and policymakers. In subsequent sections, where we discuss four principal methodologies proposed earlier, we also describe the relevant work in this regard.  

The above review showcases some of the work that has made significant contributions to positively bring the techniques of machine learning into addressing the dynamic complications of climate change. However, the ever-diversifying literature in different fields (environmental sciences, climate research, and ML/NLP) is increasingly becoming scattered, thereby reducing its effective integration with practical usage. Seeking to bridge the gap between the fields mentioned above, the present paper aims to make an action-oriented contribution to this growing body of literature by providing an integrative knowledge platform that can provide comprehensive knowledge on NLP and climate change to be used by a wide variety of actors ranging from researchers to policymakers.



\section{Media Content Analysis for Climate Change via Topic Modeling} \label{sec:lda}

Media coverage of climate change influences all strata of society from policy formulators, politicians, to general public \cite{keller2020news,carmichael2017elite, weingart2000risks,carvalho2010media,arlt2011climate,moser2010communicating}. Consequently, analyzing the content of these articles becomes important. The objective of analyzing print media content is to discover the salient themes and topics pertinent to climate change. Furthermore, the analysis would also help discover the importance of print media to the growing problem of climate change and how each of the discovered themes/topics is covered by media, and in what depth. All these analyses hold special relevance since they provide a broad yet rigorous overview of the dominant themes that emerge in the public discourse on climate change. It illuminates the interface between public policy and public discourse on climate change.   

An influential strand of literature within the environmental sciences research community has used unsupervised machine learning techniques like Latent Dirichlet Allocation (LDA) \citep{blei2003latent} to address diverse concerns for climate change raised in the media. Topic modeling using LDA has been used to analyze media discourse and trace the social policy debates on climate change in different countries like Brazil \citep{benites2018topic} and US \citep{bohr2020reporting}. \citet{lu2020evidence} apply LDA on 4875 academic articles on energy transition to uncover the dominant aspects in the literature in a meta-analytical literature review, reporting the prominence of policy-related articles along with paucity of literature from developing countries. The findings of such studies highlight the significance of these techniques in analyzing the dominant discourses and themes, yielding new insights into the various intersections that define climate policy.

\citet{keller2020news} have done an in-depth content analysis of the media coverage on climate change. Authors have used topic modeling (LDA) on $18,224$ newspaper articles published between 1997-2016 in two leading Indian English newspapers: Times of India and The Hindu. The authors used machine learning to seek answers to three broad questions, regarding, a) the salience of climate change in Indian media between 1997-2016; b) the themes and topics in media coverage on climate change, and c) the prevalence of each theme in the media coverage. The automated content analysis used by the authors extracts the latent themes from a large amount of text using the “bag-of-words” assumption. They set parameters containing 28 topics divided into five broad themes, namely, “Climate Change Impacts,” “Climate Science,” “Climate Politics,” and “Climate Change and Society” \cite[p. 20]{keller2020news}. They found the media coverage was dominated by the “climate change and society” and “climate politics” themes. The focus on the “climate change and society” theme is of particular consequence, according to the researchers, since it endeavors to educate the public and raise awareness about climate change. 


All these analyses and studies have mainly been done in the environmental science, climate, and energy research communities, where the researchers have merely used LDA as a tool to a means. However, not much work has been done in the NLP community in this regard. 

The advantage of topic modeling is that we can fetch a range of topics and prominent keywords from a large-scale corpus, including print media, social media, and policy documents, through the process of clustering. However, this process has a few limitations. We believe that researchers in the NLP community could contribute by developing advanced techniques for addressing some of these limitations. First, in topic modeling, the number of clusters or topics has to be fixed beforehand. Knowing topics in advance may not always be feasible. Matrix factorization based techniques \citep{arun2010finding} could be employed to circumvent this limitation. Second, it is not possible to determine the relationship between the topics. There has been work on learning the representations (word embedding based) for topics \cite{dieng2020topic}, this could be adapted to study the relationship between topics in the domain of climate change policy. Third, we cannot decide the evolution of topics over a period of time. NLP researchers could apply/develop dynamic topic models \citep{blei2006dynamic} that could analyze the evolution of relevant topics over time. Deep learning-based techniques could be applied to study how the language in the context of climate research has evolved over time; for example, dynamic word embeddings based techniques could be developed \citep{rudolph2018dynamic, yao2018dynamic, frermann2016bayesian}. Fourth, we can only get the topics in topic modeling but not the associated actors who have discussed the topics. For example, ‘deforestation’ can be a topic. Still, it is impossible to infer whether the ministry of environment talked about deforestation or some think-tanks or research organizations from topic modeling alone. One possible way of addressing is by the creation of a Knowledge Graph (section \ref{sec:kg}). Finally, it is impossible to find out the pro and anti-climate emotions or sentiments of various actors from topic modeling alone, however, this is crucial in policymaking. To overcome the above mentioned limitation, opinion mining techniques could be employed (section \ref{sec:sentiment}).  

\section{Opinion Mining} \label{sec:sentiment}

Mapping the overall perceptions, opinions (or sentiments), and awareness of people around specific issues can provide enlightening insights into the extent to which a social issue has penetrated the socio-political fabric of the society and how well or ill-received are the policy measures around it. Mass media, including social media and newspapers, represent spaces where such perceptions and opinions can be traced. These are the spaces where actors articulate and frame their opinions on a given issue. For example, consider these two statements from newspaper articles: \\

\noindent\fbox{\begin{minipage}{19em}

 “\textit{In India's view, the INDCs should reflect basic principles of UNFCCC and should not be legally binding.}” \cite{sibal2015}\\

“\textit{India needs to join the gathering consensus that the 2015 agreement will take the form of a legally binding treaty.}” \cite{dubash2015}.
\end{minipage}
}
\\

In the first statement, the person expresses a negative sentiment regarding the legally binding nature of any global climate agreement, especially for a country like India, which is balancing its developmental needs with combating climate change. However, in the second statement, the writer expresses a positive sentiment regarding the same notion of a legally binding treaty.

The various media spaces are replete with such opposing statements by multiple actors. Interestingly, India is among those countries where the number of tweets regarding climate change is among the highest \citep{pathak2017understanding}. NLP can provide a diverse set of tools to extract, analyze, and interpret such narratives' nature. For example, \citet{pathak2017understanding} determine the various factors that can affect the opinion of a Twitter user like the demographics (income group, age, gender), the type of affiliation (personal or organizational account), and time. In addition, studies using these techniques can also highlight the nuanced differences, like gender, that affect a person's sentiment not just towards climate change but also towards energy policies (e.g., \citep{loureiro2020sensing}.


Analysis like these can provide a constructive direction to the policy-making around an issue like climate change, where public mobilization and participation is utmost needed. It can also indicate the extent to which the existing policy measures and their outcomes converge with people's expectations, making policies more people-oriented and thereby increasing their legitimacy.

There is a plethora of work on opinion mining and sentiment analysis \cite{ravi2015survey} in the NLP domain. The majority of these approaches have been developed to adhere to commercial applications like product reviews, movie reviews, customer feedback monitoring, social media monitoring, among others. However, there has been limited application of sentiment analysis techniques for analyzing opinions about climate change policies in mass and social media. For example, \citep{jiang2017comparing} assess the public perception and sentiments regarding climate change and adaptive and mitigating measures like renewable energy. \citet{dahal2019topic} and  \citet{loureiro2020sensing} have employed sentiment analysis techniques for analyzing tweets on climate change. \citep{medimorec2015language} have assessed the difference in language used by the proponents and skeptics of climate change. 


Opinion mining in the context of climate change policy is a relatively less explored area, and more research in this direction is required. Using sentiment analysis techniques, we can identify the actors (both the person and its associated organizations) with their direct narrative, which exhibits the actors' inherent sentiment. It helps to identify which actor supports or opposes an issue. For example, the transition from coal to renewable energy may be supported by environmental focused Non-Government Organizations (NGOs) but not by the fossil fuel companies. However, sentiment analysis has its limitations. We cannot identify the actors' inherent core beliefs from their statements, which is pertinent to locating the policy subsystem. One possible way to overcome this limitation is via discourse network analysis.

\section{Beliefs and Discourse Analysis}

In climate change policies, media (including social media) is full of statements by various actors expressing different views. Analyzing such discourses can give insights into the dynamics of policymaking. The Discourse Network Analysis focuses on investigating a policy subsystem's discursive domain and emphasizes policy processes as “networked phenomenon” \citep{leifeld2017discourse}. The method is operationalized through Discourse Network Analyzer (DNA)\footnote{\url{https://github.com/leifeld/dna}} that allows mapping the policy discourse for the major actors and their beliefs along with different types of networks based on these two elements. The basic unit of analysis is the statement, which is essentially a portion of text from necessarily a public \citep{leifeld2017discourse} document (e.g., newspaper articles, policy reports) where an actor expresses a belief regarding a specific policy measure. It may be a direct quote or a paraphrased statement. This statement is coded for four main variables:
\begin{itemize}
\item the \textit{person} making the statement,
\item the \textit{organization} to which he/she belongs,
\item the \textit{concept}, which summarizes in a few words the belief being expressed, and
\item a dummy variable recording person’s \textit{agreement or disagreement }with the belief
\end{itemize}

For example, consider the following statement from a newspaper article, \\

\noindent\fbox{\begin{minipage}{19em}
“\textit{Rich nations that ``powered their way to prosperity on fossil fuel" must shoulder the greatest burden in the fight against global warming}, Prime Minister Narendra Modi wrote in the Financial Times ahead of the Paris climate talks on Monday.” \cite{ht2015}
\end{minipage}
}
\\

DNA would code this as (a) Narendra Modi [the \textit{person}], (b) Prime Minister’s Office [the \textit{organization}], (c) the concept category “Developed countries should bear the major responsibility in Greenhouse Gas reduction” [the \textit{concept}], and (d) ‘agreement’ with the concept [dummy variable for \textit{agreement/disagreement}]. Consider another example of a direct quote from another newspaper article, \\

\noindent\fbox{\begin{minipage}{19em}
“\textit{Well, what I mean by saying that this is not Kyoto is......What was a distinguishing feature of Kyoto is that all of the new obligations were only directed at developed countries, so it was really a developed country agreement with developing countries on the sidelines. That doesn’t work anymore.},” \cite{hindu2016}
\end{minipage}
}
\\

This statement is coded as, (a) Todd Stern [the \textit{person}] (b) United States Government [the \textit{organization}], (c) concept category “Developed countries should bear the major responsibility in Greenhouse gas reduction” [the \textit{concept}], and (d) ‘disagreement’ with the concept [dummy variable for \textit{agreement/disagreement}].

Following this procedure, once the statements from the documents are coded for the variables, the data can be extracted to derive various types of networks for analysis. These networks can be analyzed using popular network visualization software like VISONE \cite{brandes2004analysis}, allowing for cluster analysis. This type of analysis automatically groups those actors who are most densely connected based on ideological affinity. 

In a comparative study including U.S., Canada, Brazil, and India, \citet{kukkonen2018international} have employed discourse network analysis technique to arrive at an understanding of climate change discourse and dominant actors within it. They specifically look at how influential are the international actors within national discourse. They find meaningful actor clusters, theoretically contextualized as advocacy coalitions in the U.S. and Canada while uncovering none in Brazil and India. However, they find that the international actors are more dominant in low-income countries like Brazil and India, where the climate change debate is less polarized. Such studies have a direct bearing on our understanding of climate policy-making. The emerging actor-networks defending or opposing particular policy norms highlight the advocacy patterns within a policy subsystem and the future direction that it may take. \citet{veltri2017climate} have done a thematic and network analysis of the Twitter data relevant to climate change showing how a small set of media sources play a significant role in sharing climate-related information. 


\begin{figure*}[t]
\centering
\includegraphics[width=\linewidth]{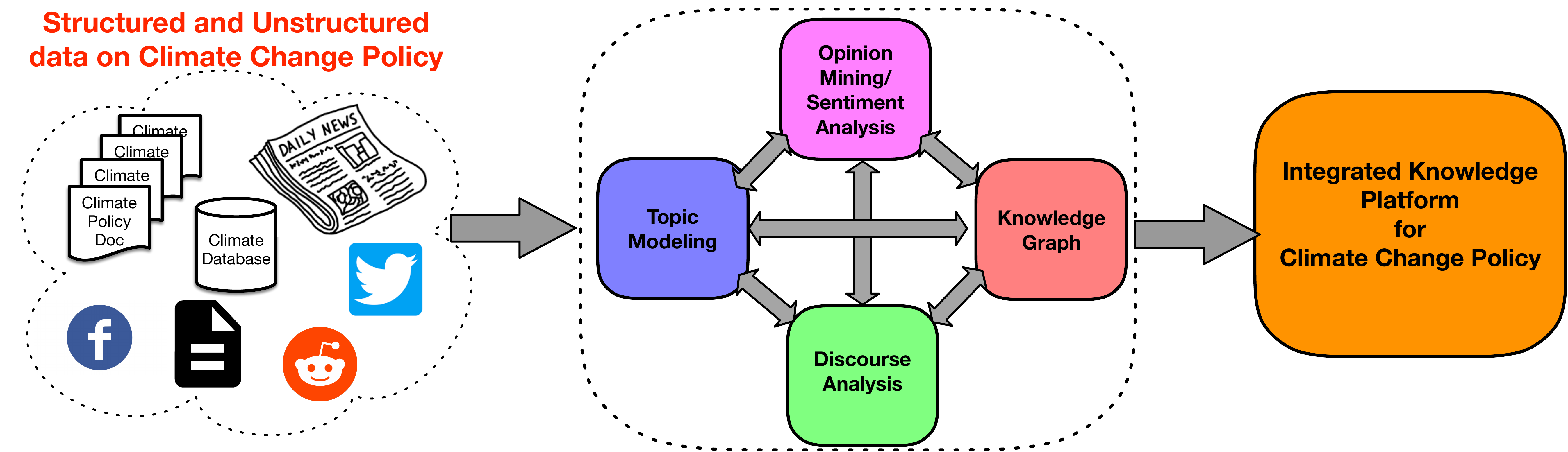}
\caption{Integrated Knowledge Platform for Climate Change Policy}
\label{fig:knowledgePlatform}
\end{figure*}

The method highlights the significance of the discourse analysis for policy-making dynamics. Conceiving the public discourse as a space for debating actors and connecting them according to similar and conflicting beliefs allows for analyzing the ideological affinities and antagonisms. This provides a glimpse into the policy trajectories that may emerge within a particular policy subsystem like climate change by analyzing the existing and possible advocacy coalitions. In the NLP community, there has been much research in the context of identifying entities and relationships between them in text via the tasks of Named Entity Recognition (NER) \citep{wen2019survey, li2020survey}, Relationship Extraction \citep{pawar2017relation} and Open Information Extraction \citep{niklaus-etal-2018-survey}. Many of these techniques could be further developed for analyzing discourse in media text in the vein of discourse network analysis. 


Discourse analysis gives us a range of actors and their similarities or differences in beliefs. This method already exists, and researchers have conducted it through discourse network analysis. However, hand-coding in DNA has some limitations in terms of scale. However, using NLP techniques, we can train the system on hand-coded data and code more information in a large text corpus. However, the discourse network analysis has its limitation. It deals with two things: actors and their beliefs. However, an actor might have multiple attributes (based on social status and role) relevant to its climate opinion. Discourse network analysis fails to capture the multi-view of an actor. To address this, we propose creating a climate knowledge graph.

\section{Climate Knowledge Graph} \label{sec:kg}

Much research has been done in the NLP community concerning knowledge graphs \citep{ji2020survey}. A Knowledge Graph (KG) is a structured representation of information about the world. KG represents this information in the form of a graph where nodes are the entities, and edges indicate the relationships between them. The notion of entities is quite generic. An entity could be a specific person or an object or a concept, or an organization. Several KGs have been created (e.g.,YAGO \citep{suchanek2007yago}, DBPedia \citep{auer2007dbpedia}, FreeBase \cite{bollacker2008freebase}, NELL \cite{carlson2010toward}, WikiData \citep{vrandevcic2014wikidata}, inter alia). However, these KGs capture the general factual knowledge about the world and are not specific to climate change policy. We want to represent unstructured knowledge about climate change policy in the form of a KG where entities would be climate policy actors, politicians, activists, organizations, and concepts related to climate change. Edges will represent different types of relationships between entities, e.g., beliefs of entities about other entities and an entity's association with another entity. KGs are created from an unstructured text by extracting domain-related entities and relations between them. For creating a Climate KG, we need to create a domain-specific dictionary, i.e., climate dictionary. The climate dictionary would focus only on terms (entities and relations) of interest, e.g., it could have as an entity, a politician and climate-related organizations/forums they are associated with, and relationships could be beliefs expressed by them at different forums. 

Various Governments are a producer of a wealth of information on policy decisions. For coming up with policies, Governments consults a range sources from research outputs to media reports \citep{wang2019government}. In climate change, a policy decision is complex because it combines science, politics, economics, and social aspects. Information comes from various sources and attributes. Moreover, as described in the case of sentiment and discourse analysis, the relationship between the actor and their attribute is crucial to making a rational decision. Climate specific KG can help to overcome this gap.

\section{Knowledge Platform: Connecting the Dots}
We envisage an integrated knowledge platform in the form of a database. Given structured and unstructured data in the form of print and social media, government reports, white papers produced by NGOs related to climate change policy, we could employ the four methodologies described earlier to create a database (Figure \ref{fig:knowledgePlatform}). The exact nature of the database still needs more research, but one possibility could be graph databases. 

The platform will serve as both a research tool facilitating analysis of large volumes of unstructured text-based data and a knowledge and action network. Designing the final platform will ensure ease of access and utilization by a wide range of decision-makers. The practice of sustainable development concerning climate change policy is generally undertaken by large financial, scientific, and governmental institutions within an organizational framework that involves a limited set of domain expertise. Moreover, the actual decision-maker has to trust the experts' knowledge without having an idea of the big picture. Since the policymakers are not expected to have a technical background, whatever is the actual implementation of the knowledge platform, it must support natural language queries. Anyone should be able to ask a question in natural language about various policies, major actors, their opinions, and the relationship between them. Recently, there has been research in this direction in the NLP domain \citep{dar2019frameworks, kim2020natural}, those techniques could be further developed for our case.   

Our platform will evolve as a solution to these limitations through four significant steps. By topic modeling, we expect to get the dominant topics of the discussions. With opinion mining and discourse analysis, we would like to identify the actors in the climate debate and their relationship based on similar and different beliefs. For example, actor A and actor B  may support nuclear energy, but they might not have interacted in reality. By identifying their perception, beliefs, and sentiments, we can create a cluster of actors. Finally, we make an integrated system covering a range of attributes of the actors through a knowledge graph. 

To give an example, how all the discussed use-cases could help create a Knowledge Platform, we take India's use-case. The rationale being that India represents a paradox in terms of climate change concerns. With the second largest population globally, it is the third highest emitter globally \citep{wang2020modeling} but owing to the deep-rooted socio-economic gap; its per-capita emissions are much below the global average \citep{anandarajah2014india}. In this regard, the media representation of climate change policy-related news in prominent English national newspapers in India could be investigated. Climate-related media articles could be extracted from the Factiva database\footnote{\url{https://professional.dowjones.com/factiva/}}. Making use of techniques described in-depth previously, we could extract the reported speeches, both direct and indirect, from the media along with the actors and organization associated with them. Furthermore, these speeches could be classified based on pre-defined beliefs. Finally, a knowledge graph could be created having actors as nodes and edges as beliefs. In the end, we would get a knowledge platform that could be used by Indian climate policymakers.

\section{Conclusion}

Climate change is a reality, and it needs to be addressed on a priority basis. Formulating climate change policies is the first step towards it. However, to make informed decisions, policymakers need to take into account all sources of information. We showed in the paper climate change researchers have mostly been working in isolation analyzing media discourses, and on the other side, NLP researchers are completely oblivion to the research questions in the domain of climate change. In this paper, we argued that NLP researchers and Climate change researchers could benefit from each other and contribute to the greater good of society. 

This paper advocated creating an integrated knowledge platform that would extract relevant information coming from structured and unstructured sources and store it in a suitable database. A policymaker could query the knowledge platform in natural language and ask about policies, opinions, and actors' beliefs. We outlined four methodologies for creating the knowledge platform: topic modeling, opinion mining, discourse network analysis, and knowledge graphs. NLP researchers have a rich experience in these four, and they could, in conjunction with climate researchers, contribute actively towards creating the knowledge platform. Although this paper discussed only four methods, this is not an exclusive list, and many more avenues are possible. Active collaboration between the two research communities (NLP and climate change) would open up unexplored avenues leading to good for humanity.

\bibliographystyle{acl_natbib}
\bibliography{acl2021}


\end{document}